\pdfoutput=1
\documentclass{article}

\PassOptionsToPackage{numbers, compress}{natbib}

\usepackage[preprint]{neurips_2023}

\usepackage[utf8]{inputenc} 
\usepackage[T1]{fontenc}    
\usepackage{hyperref}       
\usepackage{url}            
\usepackage{booktabs}       
\usepackage{amsfonts}       
\usepackage{nicefrac}       
\usepackage{microtype}      
\usepackage{xcolor}         

\usepackage{amsmath}
\usepackage{multirow}
\usepackage{tikz}
\usepackage{color}
\usepackage{pifont}
\usepackage{amssymb}
\usepackage{enumitem}
\usepackage{xspace}

\usepackage{graphicx}
\usepackage{caption} 
\usepackage{subcaption}
\usepackage{wrapfig}
\usepackage{graphicx}
\usepackage{amsmath}
\usepackage{amssymb}
\usepackage{booktabs}
\usepackage{adjustbox}
\usepackage{multirow}
\usepackage{arydshln}
\usepackage{colortbl}
\usepackage{amsfonts}       
\usepackage{nicefrac}       
\usepackage{microtype}      
\usepackage{times}
\usepackage{epsfig}
\usepackage{tipa}
\usepackage[T1, OT1]{fontenc}
\DeclareTextSymbolDefault{\dh}{T1}
\usepackage{booktabs}
\usepackage{multirow}
\usepackage{color, colortbl}
\usepackage{pifont}
\usepackage{makecell}
\usepackage{threeparttable}
\usepackage{xcolor}
\usepackage{scalerel}
\usepackage{makecell}
\usepackage{tabu}
\usepackage{tikz}
\usepackage{marvosym}
\usepackage{xcolor} 

\definecolor{Gray}{gray}{0.5}
\definecolor{LGray}{gray}{0.9}
\definecolor{darkblue}{RGB}{94,110,186}
\definecolor{darkGreen}{RGB}{92, 148, 110}
\definecolor{myblue}{RGB}{14, 121, 178}
\definecolor{rightpath}{RGB}{248, 203, 173}
\definecolor{wrongpath}{RGB}{89, 89, 89}

\usepackage{tabulary}
\usepackage{tabulary,multirow,overpic,xcolor}
\newcolumntype{x}[1]{>{\centering\arraybackslash}m{#1pt}}
\newcolumntype{y}[1]{>{\arraybackslash}m{#1pt}}

\title{R1-Track: Direct Application of MLLMs to Visual Object Tracking via Reinforcement Learning}

\author{Biao Wang$^{1,\dagger}$, Wenwen Li$^{1}$, and Jiawei Ge$^2$  \\
    \small$^1$Beihang University~~~~\small$^2$Southeast University  \\
    \url{wangbiao301@buaa.edu.cn}~~~~\url{jiawei_ge@seu.edu.cn}\\
    {\small \url{https://github.com/Wangbiao2/R1-Track}} \\
}

\begin{document}

\maketitle
{
\renewcommand{\thefootnote}%
{\fnsymbol{footnote}}
\footnotetext[0]{$\dagger$ Corresponding author.} 
}

\begin{abstract}
Visual single object tracking aims to continuously localize and estimate the scale of a target in subsequent video frames, given only its initial state in the first frame. This task has traditionally been framed as a template matching problem, progressing through correlation filters, two-stream networks, and
one-stream networks. However, these methods typically require explicit classification
and regression modeling, depend on supervised training on large datasets, and
lack flexibility due to their focus on tracking. 
In recent years, multi-modal large language models (MLLMs) have advanced rapidly. Open-source models such as Qwen2.5-VL, a flagship MLLMs with strong foundational capabilities, demonstrate excellent performance in grounding tasks. This has spurred interest in applying such models directly to visual tracking. However, experiments reveal that Qwen2.5-VL struggles with template matching between image pairs (i.e., tracking tasks). Inspired by DeepSeek-R1, we fine-tuned Qwen2.5-VL using group
relative policy optimization (GRPO) reinforcement learning on a small dataset
with a rule-based reward function. The resulting model, R1-Track, achieved notable performance on the GOT-10k benchmark. R1-Track supports flexible initialization via bounding boxes or text descriptions while retaining most of the original model's general capabilities. We further discuss potential improvements for R1-Track. This rough technical report summarizes our findings as of May 2025.
\end{abstract}

\section{Introduction}
\label{sec:Introduction}

Visual object tracking is a fundamental task in computer vision and robotics, where the goal is to continuously estimate the location and scale of a target object across video frames given only its initial state in the first frame. Most existing tracking methods primarily rely on template matching frameworks, which extract the target region from the initial frame as a template and perform similarity comparisons within search regions of subsequent frames. The field has evolved through three major paradigms: correlation filtering~\cite{KCF,ECO}, dual-stream siamese networks~\cite{SiamRPN++, TransT}, and single-stream ViT architectures~\cite{OSTrack}. Notably, deep learning-based approaches in this domain typically require explicit classification and regression modeling with supervised training on large-scale annotated datasets. However, these models often incorporate heuristic design choices that limit their flexibility and hinder generalization to diverse tasks.

Since the release of GPT-3~\cite{GPT3}, large language models (LLMs) have attracted significant attention and experienced rapid development. Alongside this progress, multimodal large language models (MLLMs) have also advanced considerably, leading to the emergence of high-performing open-source models such as LLaVA~\cite{LLaVA}, InternVL~\cite{internvl}, and Qwen2.5-VL~\cite{qwen2.5vl}. These models can process multiple modalities, including text,
images, and videos, and perform downstream tasks such as object detection,
OCR, and mathematical reasoning.

This naturally raises the following question: \textbf{\textit{Can MLLMs be directly applied to visual object tracking tasks?}}

Unfortunately, our experiments show that Qwen2.5-VL is not effective for target tracking. Specifically, even with extensive prompt engineering, the native
Qwen2.5-VL model fails to relocate a specified target across image pairs.

\textbf{\textit{So, how can this issue be effectively addressed?}}

Supervised fine-tuning (SFT) of MLLMs is a straightforward but
potentially suboptimal approach. First, SFT risks catastrophic forgetting, weakening the model's
general capabilities. Secondly, SFT requires hard supervision for every coordinate point (e.g., $[x_{min},y_{min},x_{max},y_{max}]$), making it less suitable for fine-grained grounding tasks. Recent advancements in applying reinforcement learning (RL) to
LLMs have shown remarkable progress. As evidenced by O1~\cite{openai-o1}, the implementation of test-time scaling strategies has shown substantial potential in enhancing LLMs' capacity for complex reasoning. Subsequently, DeepSeek-R1-Zero~\cite{deepseekr1} revealed that
strategically applying a rule-based reward system for reward modeling can
effectively leverage RL to unlock exceptional reasoning and cognitive
capabilities in LLMs, even without extensive SFT.

Accordingly, we also adopt a RL framework to fine-tune MLLMs, employing the GIoU metric as a soft reward function to guide parameter updates. We explore both ``\textit{think}'' and ``\textit{no-think}'' training strategies. Aligned with Deepseek-R1, we implement group relative policy optimization (GRPO)~\cite{grpo} as the RL optimization algorithm, naming the resulting model \textbf{R1-Track}.

R1-Track achieves superior performance in visual object tracking tasks, attaining an average overlap (AO) score of 0.68 on the GOT10k~\cite{GOT10k} benchmark. Our contributions are summarized as follows:

\begin{itemize}
\item We investigate the direct application of MLLMs to visual object
tracking, constructing specialized datasets by sampling the GOT10k training
set;

\item We implement SFT and RL strategies to fine-tune MLLMs for acquiring tracking capabilities while supporting flexible initialization protocols;

\item We conduct comprehensive experiments, make our code, datasets, and
model weights publicly available, and discuss future research directions.
\end{itemize}

\section{Data}
\label{sec:Data}
Here, we introduce the datasets used for R1-Track training and
testing. For training, we sampled R1-Track-5k, R1-Track-100k, and cold-start
data. The test data is the official GOT-10k benchmark.
 
\paragraph{R1-Track-5k.}
R1-Track-5k is a simple dataset designed for preliminary experimental validation, comprising 5,000 template-search image pairs. We randomly selected 5,000 videos from the GOT-10k training set and extracted two random frames from each video. Centering on the target object in each frame, we randomly cropped
the images and resized the cropped regions to $336 \times 336$ pixels. It should be noted that R1-Track-5k has a significant limitation: all target objects in the resized images are forced to a 1:1 aspect ratio.

\begin{figure*}[t]
	\centering
	\includegraphics[width=0.75 \textwidth]{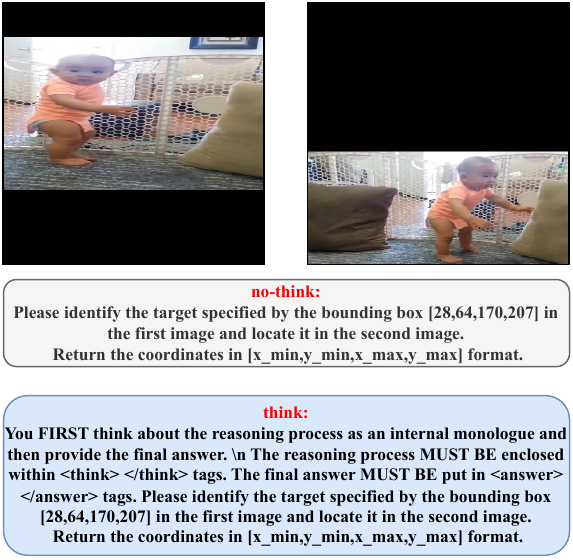} 
	\caption{
		Visualization of R1-Track-5k dataset. 
	}
	\label{fig2:R1-Track-5k}
\end{figure*}

Figure~{\ref{fig2:R1-Track-5k}} presents the visualization of R1-Track-5k. Prompts under the ``think'' mode impose additional formatting requirements compared to those in the ``no-think'' mode.

\paragraph{R1-Track-100k.}
R1-Track-100k is a larger-scale dataset containing 100,000 template-search image pairs, covering all videos in the GOT-10k training dataset. It supports multiple input resolutions (112×112, 224×224, 336×336, and 448×448). During sampling, we set the search region scale factor to 2-8, center shift factor to 0-0.2, and frame intervals ranging from 1 to the total number of video frames. The task prompt was updated to: 
\begin{verbatim}
Given two images, you need to:
1. Analyze and identify the target object marked by bounding box 
   <BBOXFLAG> in <image_1>;
2. Re-locate this target in <image_2>;
3. Return [x_min, y_min, x_max, y_max] coordinates of the target 
   in <image_2>.
\end{verbatim}
The formatting prompt remains consistent with R1-Track-5k.

\paragraph{Cold-Start Data.}
The SFT and RL fine-tuning processes use
identical datasets; differences exist solely in their data assembly formats.

Before SFT or RL, we performed a cold start for the base model,
also achieved through SFT. The cold-start dataset consists of 100 ``no-think'' data points
and 10 ``think'' data points to equip the model with basic tracking
capabilities and accelerate training. All subsequent procedures involved further fine-tuning of the
cold-start model. In other words, the training workflow follows the path of ``base model $-->$ cold start $-->$ SFT/RL".

\section{Methodology}
\label{sec:Methodology}

We first briefly review the Group Relative Policy Optimization (GRPO)~\cite{grpo}. Then, we demonstrate how we design the tracking rewards for GRPO to enhance MLLMs. Finally, the architecture and implementation details of the R1-Track tracker are introduced.

\subsection{Group Relative Policy Optimization}
\label{subsec:Preliminary}

Group Relative Policy Optimization (GRPO)~\cite{grpo} is a variant of Proximal Policy Optimization (PPO)~\cite{ppo} in reinforcement learning. By directly comparing groups of candidate responses, GRPO
eliminates dependence on a critic model and significantly lowers training
resource requirements.
Given an input question $q$, GRPO first generates $g$ distinct candidate responses $o=\{o_1, \dots,o_g\}$ through policy sampling.
The MLLM serves as the reward function to obtain the corresponding scores $\{r_1, \dots, r_g\}$.

 GRPO computes their mean and standard deviation for normalization
to determine the quality of these responses:
\begin{equation}
    A_i=
    \frac{r_i-\mathrm{mean}(\{r_i\}_{i=1}^g)}{\mathrm{std}(\{r_i\}_{i=1}^h)} \text{,}
\end{equation}
where $A_i$ represents the relative quality of the $i$-th answer. GRPO encourages the model to favor better answers with higher
scores within the group.
The final training objective also considers preventing the optimized policy $\pi_\theta$ from deviating far from the base MLLM (step 0) parameters $\pi_\mathrm{ref}$ by adding a KL-divergence term $\mathrm{D}_\mathrm{KL}$:
\begin{equation}
    \max_{\pi_\theta} \mathbb{E}_{o\sim \pi_{\theta_{\mathrm{old}}}(p)} \Big[
         \sum_{i=1}^g \frac{\pi_\theta (o_i)}{\pi_{\theta_{\mathrm{old}}}(o_i)} \cdot A_i  - \beta\, \mathrm{D}_\mathrm{KL}\Big(\pi_\theta \,\Vert\, \pi_\mathrm{ref}\Big)
    \Big] \text{,}
\end{equation}
where $\beta$ is a regularization coefficient, preventing excessive deviation from the reference policy during optimization.

R1-Track's training framework adopts EasyR1~\cite{easyr1},
employing a token-level policy gradient loss similar to DAPO~\cite{dapo}.


\subsection{Tracking Rewards}

We explore how to use GRPO to enhance the performance of MLLMs in visual object tracking. The overall reward comprises three components: format, answer, and
length. Reward function design varies slightly depending on whether the model
incorporates reasoning.
\begin{equation}
	R_{\mathrm{overall}} = a \times R_{\mathrm{answer}} + b \times R_{\mathrm{format}} + c \times R_{\mathrm{length}}
\end{equation}

\paragraph{Format reward.} To enable the model to output responses in the desired format. In ``no-think'' mode, we expect the model to directly output
standard bounding box coordinates as \texttt{$[x_{min}, x_{max}, y_{min}, y_{max}]$}.
In ``think'' mode, the model should generate outputs in the
following format: \texttt{<think>...</think><answer>$[x_{min}, x_{max}, y_{min}, y_{max}]$</answer>}. We use regular expression matching to determine whether the model
adheres to the specified format:

\begin{equation}
\begin{cases}
R_{\mathrm{format}} = 1.0, & \text{if output matches format}, \\
R_{\mathrm{overall}} = -1.0, & \text{if output doesn't match format}.
\end{cases}
\end{equation}

\paragraph{Answer reward} Consistent with most tracking methods, we use GIoU~\cite{GIoU} to measure the overlap between the predicted and ground-truth bounding boxes. The answer reward is a piecewise function with GIoU as its independent variable. 

\begin{equation}
	R_{\mathrm{answer}} = \begin{cases}
		\text{GIoU}, & \text{if  } \text{GIoU} <= 0, \\
		0, & \text{if  } 0 < \text{GIoU} <= 0.4, \\
		\text{GIoU}, & \text{if  } 0.4 < \text{GIoU} <= 0.75, \\
		\text{GIoU}+0.2, & \text{if  } 0.75 < \text{GIoU} <= 0.95, \\
		\text{GIoU}+0.5, & \text{if  } \text{GIoU} > 0.95.
	\end{cases}
\end{equation}

\paragraph{Length reward}
For the ``think'' mode, we introduce an additional length reward to prevent responses that are either too long or too short. It is defined as follows:
\begin{equation}
	\begin{cases}
		R_{\mathrm{overall}} = -1.0, & \text{if } L <= L_{min}, \\
		R_{\mathrm{length}} = 0, & \text{if } L_{min}<L<=L_{cache}, \\
		R_{\mathrm{length}} = \frac{L_{cache}-L}{L_{max}-L{cache}}, & \text{if } L_{cache}<L<=L_{max}. \\
	\end{cases}
\end{equation}

\subsection{R1-Track}
The overall framework of R1-Track is illustrated in Figure~\ref{fig3:R1-Track-framework}.

Considering both performance and speed, we selected the lightweight Qwen2.5-VL-3B-Instruct as the base model for fine-tuning R1-Track. First, the model is fine-tuned on the cold-start data using a relatively low learning rate. Subsequently, more refined SFT or RL fine-tuning is performed. Notably, jointly fine-tuning the vision encoder and the LLM significantly improves performance. More detailed parameter settings can be found in our provided code repository.

During inference, we recommend deploying the service using vLLM~\cite{vllm} and invoking it on demand for tracking tasks. Following the one-shot principle of conventional trackers, R1-Track assumes the target is provided as a bounding box in the first frame. Based on this initial input, a template is cropped and cached. The model then performs tracking frame by frame through images understanding, reasoning, and localization. Throughout the entire process, the template remains fixed, and no target re-localization strategies are applied.

When only a textual description of the target is provided in the first frame, R1-Track can first obtain the corresponding bounding box through a question-answering mechanism, and then proceed with the tracking task following the same steps as described above. We set the prompt for this scenario as:

\begin{verbatim}
	Please return the coordinates of {text_description} in JSON format.
\end{verbatim}

\begin{figure*}[t]
	\centering
	\includegraphics[width=1 \textwidth]{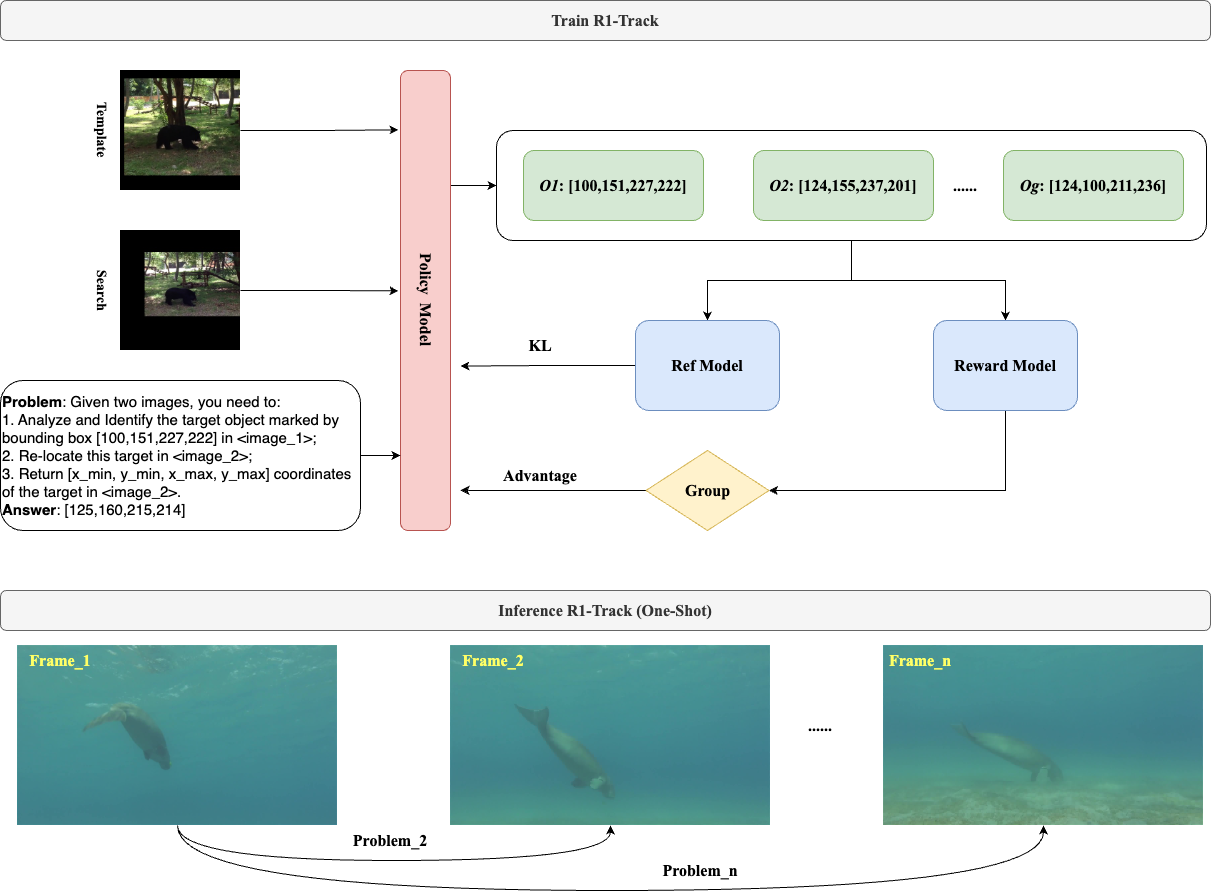} 
	\caption{
		The overall framework of R1-Track. 
	}
	\label{fig3:R1-Track-framework}
\end{figure*}

\section{Experiments}
\label{sec:Experiments}

We conducted evaluations only on the GOT-10k benchmark. To comply with its testing protocol, all our training data were sourced exclusively from the GOT-10k training set. Based on the dataset used, the training algorithm, and whether the ``think'' mode was enabled, we developed a total of six models and evaluated their performance individually, as summarized in Table~\ref{tab:tracking}. It is worth noting that during inference, both the template and search images are resized to $336\times336$ pixels before being fed into R1-Track along with the prompt. A video demonstration of the tracking results is available at \url{https://www.youtube.com/watch?v=jJUT1lQHYEE}.

	\begin{table}[htbp]
		\centering
		\renewcommand{\arraystretch}{1.2}
		\setlength{\tabcolsep}{6pt}
		\begin{tabular}{>{\raggedright\arraybackslash}p{2.5cm} *{7}{c}}
			\toprule
			 &  \textbf{Finetune Data} & \textbf{Think?}& \textbf{AO} & \textbf{SR$_{0.5}$} & \textbf{SR$_{0.75}$} & \textbf{Params}  \\
			\midrule
			LLaVA-1.5~\cite{llava-1.5} & -  & No & - & 0.235 & 0.202 & 7B  \\
			Qwen2.5-VL-7B-Instruct~\cite{qwen2.5vl} & - & No & - & 0.126 & 0.011 & 7B  \\
			\hline

			\addlinespace
			R1-Track-SFT &  R1-Track-5k & No & 0.543 & 0.633 & 0.338 & 3B  \\
			R1-Track-GRPO &  R1-Track-5k & Yes & 0.586 & 0.676 & 0.470 & 3B  \\
			R1-Track-GRPO &  R1-Track-5k & No & 0.585 & 0.673 & 0.500 & 3B \\
			\addlinespace
			\hline
			R1-Track-SFT &  R1-Track-100k & No & 0.667 & 0.746 & 0.620 & 3B \\
			R1-Track-GRPO & R1-Track-100k & Yes & 0.672 & 0.759 & 0.624 & 3B  \\
			R1-Track-GRPO &  R1-Track-100k & No & 0.680 & 0.766 & 0.637 & 3B \\

			\addlinespace
			\hline
			TransT~\cite{TransT} & GOT-10k * & - & 0.671 & 0.768 & 0.609 & 23 M  \\
			OSTrack~\cite{OSTrack} &  GOT-10k * & - & 0.710 & 0.804 & 0.682 & 92 M \\
			\bottomrule

		\end{tabular}
				\caption{\textbf{Results on GOT-10k.} The * symbol indicates that the training data size for these tracking methods is typically more than 50 times that of R1-Track-100k.}
		\label{tab:tracking}
	\end{table}

As shown in Table~\ref{tab:tracking}, the original Qwen2.5-VL model performs poorly on the tracking task. However, fine-tuning on merely 5k low-quality samples via SFT leads to a noticeable performance improvement. When scaling up the training data size, the AO score on the GOT-10k benchmark improves by an additional 12\%. The results also show that fine-tuning MLLM with GRPO and a rule-based reward function yields better performance than SFT.

Interestingly, we find that letting the 3B base model directly output the results instead of following the \texttt{<think></think><answer></answer>} format leads to higher scores on GOT-10k. Potential reasons for this phenomenon include insufficient model capacity, limited or low-quality cold-start data, or unsuitability of the reasoning model for tracking tasks.

Moreover, we observe that R1-Track achieves comparable performance to other expert trackers trained on much larger datasets, despite using significantly less fine-tuning data.

\section{Conclusions and Discussions}
\label{sec:Conclusions}

In this work, we explore the direct application of MLLMs for visual object tracking through fine-tuning. While both SFT and RL are effective, training with GRPO and rule-based reward functions shows greater improvement in tracking performance and causes less damage to the original model. Although R1-Track performs well on the GOT-10k benchmark, it still lags behind state-of-the-art expert models. Nonetheless, our findings suggest that future work may not require training dedicated expert trackers; instead, incorporating tracking-related data during the pre-training or post-training stages of MLLMs may be sufficient to enable effective tracking performance.

In the following, we discuss potential directions for improving R1-Track:

\begin{itemize}
\item Generating higher-quality cold-start tracking data with reasoning traces;  
\item Exploring larger base models, such as those with 7B or 72B parameters;  
\item Enhancing temporal modeling (e.g., multi-frame or video-based approaches) to improve sequential discrimination, which currently limits performance;  
\item Expanding the model's capabilities to support various multimodal tracking tasks, such as RGB, RGB-T, RGB-E, and RGB-D through additional training data;  
\item Exploring performance improvements via tool calls or image-based reasoning methods like ``thinking with images''.
\end{itemize}

\clearpage


\end{document}